\documentclass[10pt,twocolumn,letterpaper]{article}

\usepackage{iccv}
\usepackage{times}
\usepackage{epsfig}
\usepackage{graphicx}
\usepackage{amsmath}
\usepackage{amssymb}

\makeatletter
\newcommand{\thickhline}{%
    \noalign {\ifnum 0=`}\fi \hrule height 1pt
    \futurelet \reserved@a \@xhline
}
\makeatother
\usepackage[pagebackref=true,breaklinks=true,letterpaper=true,colorlinks,bookmarks=false]{hyperref}

\iccvfinalcopy 

\def\iccvPaperID{6166} 

\ificcvfinal\fi
\begin{document}
\onecolumn

\noindent This paper was the result of a master thesis at ETH Zurich. Personal use of this material is permitted. Permission from the authors must be obtained for all other uses, in any
current or future media, including printing/publishing this material for advertising or promotional purposes, creating
new collective works, for sale or distribution to servers or lists, or use of any copyrighted component of this work in
other works.\\

\noindent Please cite this paper as:\\

{\fontfamily{qcr}\selectfont \noindent@masterthesis\{haldimann2019dissim,

\begin{tabular}{lll}
\begin{tabular}[c]{@{}l@{}} title \\ \  \end{tabular} & \begin{tabular}[c]{@{}l@{}} = \\ \  \end{tabular} & \begin{tabular}[c]{@{}l@{}}"This is not what I imagined: Error Detection for Semantic Segmentation \\ 
through Visual Dissimilarity",\end{tabular} \\
author & = & "Haldimann, David and Blum, Hermann and Siegwart, Roland and Cadena, Cesar",              \\
school & = & "ETH Zurich",           \\
month   & = & "March", \\
year   & = & "2019"                                                            
\end{tabular}

\noindent \}

}

\thispagestyle{plain}
\newpage
\twocolumn
 \title{This is not what I imagined:\\
 Error Detection for Semantic Segmentation through Visual Dissimilarity}

\author{David Haldimann\\
ETH Zurich\\
{\tt\small haldimann.d@gmail.com}
\and
Hermann Blum\\
ETH Zurich\\
{\tt\small blumh@ethz.ch}
\and
Roland Siegwart\\
ETH Zurich\\
{\tt\small rsiegwart@ethz.ch}
\and
Cesar Cadena\\
ETH Zurich\\
{\tt\small cesarc@ethz.ch}
}

\maketitle

\begin{abstract}
There has been a remarkable progress in the accuracy of semantic segmentation due to the capabilities of deep learning.
Unfortunately, these methods are not able to generalize much further than the distribution of their training data and fail to handle out-of-distribution classes appropriately. This limits the applicability to autonomous or safety critical systems.
We propose a novel method leveraging generative models to detect wrongly segmented or out-of-distribution instances. Conditioned on the predicted semantic segmentation, an RGB image is generated. We then learn a dissimilarity metric that compares the generated image with the original input and detects inconsistencies introduced by the semantic segmentation.
We present test cases for outlier and misclassification detection and evaluate our method qualitatively and quantitatively on multiple datasets.
   
\end{abstract}

\section{Introduction}
Semantic segmentation~\cite{long2015fully,valada2017adapnet,chen2018deeplab,chen2018encoder} has seen significant advances with state-of-the-art methods precisely segmenting and classifying high resolution images. However, all best performing methods are based on deep learning, which is known to be overconfident and inaccurate outside of the training distribution. In vision, such failure cases are not only adversarial examples generated by security researchers, but include very different weather and lighting conditions or unknown object classes. For robust and reliable deployment in real-world conditions, a semantic segmentation model therefore needs to be able to detect input that differs from the training set. This task is known as novelty or anomaly detection. Standard approaches in novelty detection allow to classify the whole input as in or out-of-distribution (OoD). For semantic segmentation, we require a method that can detect local OoD instances in parts of the input.

\begin{center}
 \begin{figure}
 \includegraphics[width=1.0\linewidth]{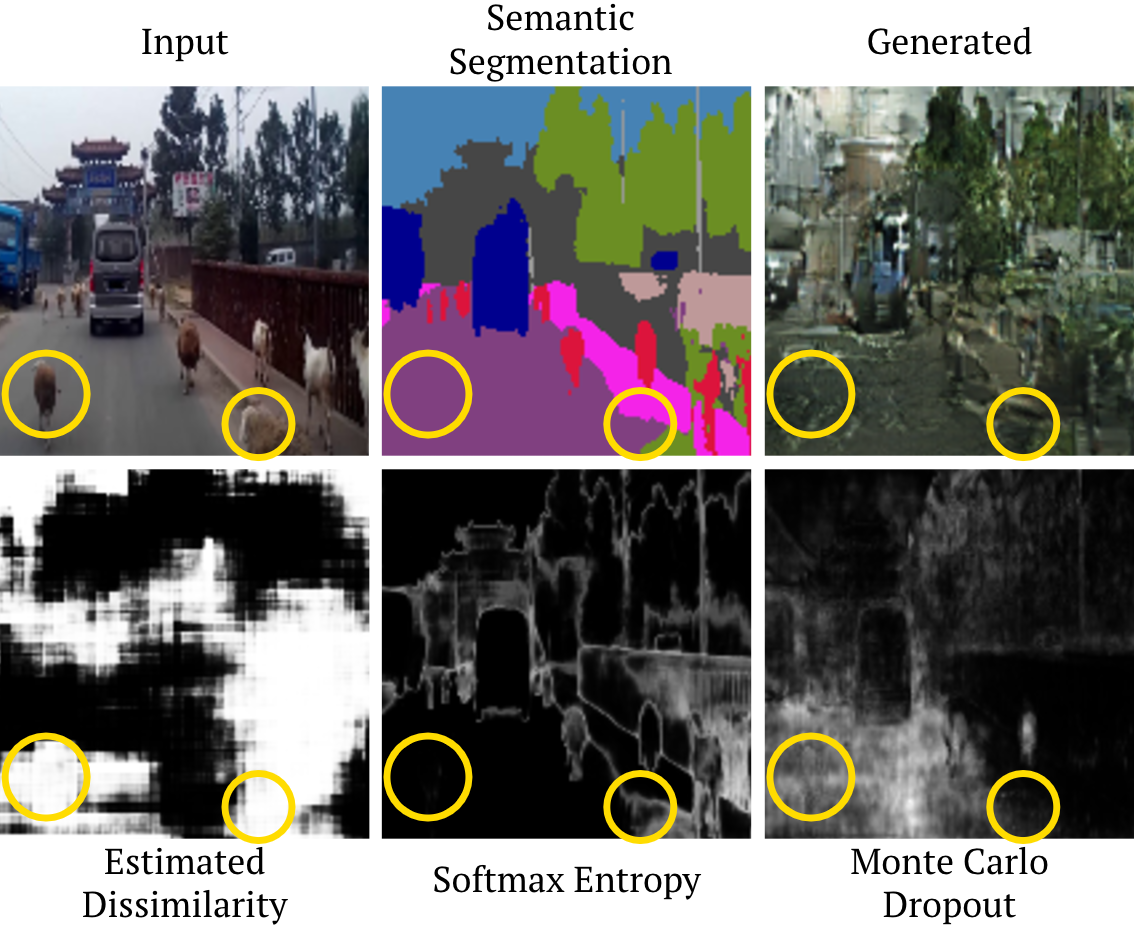}
 \caption{Generative capabilities of adversarial networks are leveraged to detect erroneous classifications by learning a dissimilarity metric.}
 \label{fig:teaser}
 \end{figure}
\end{center}
In this work, we utilize generative models such as the conditional generative adversarial network (cGAN)~\cite{pix2pix2016} to generate synthetic images based on the output of a semantic segmentation network. The synthetic images are compared with the original images in order to detect objects missed or wrongly classified by the semantic segmentation. Unfortunately, usual reconstruction metrics like the difference to the original image have no meaning for this case, as e.g. a car might be reconstructed in a different color than the original car. This problem is closely related to problems of measuring the `realism' of GAN outputs. Similar to perceptual losses~\cite{johnson2016perceptual}, we compare the images with a convolutional neural network that estimates a learned dissimilarity between real and synthetic images. Based on this output misclassified or OoD instances are identified locally.
To summarize, we propose a novel detection method for semantic segmentation networks based on dissimilarity to distinguish wrongly classified and OoD objects. This kind of feature allows any semantic segmentation approach to be extended to detect OoD instances and further utilize this dissimilarity output for higher-order decision networks. To our knowledge, visual dissimilarity for local novelty detection has not been examined in literature so far.

We test our proposed method on different datasets for both OoD and misclassification detection.
\begin{figure*}
\begin{center}
\includegraphics[width=0.8\linewidth]{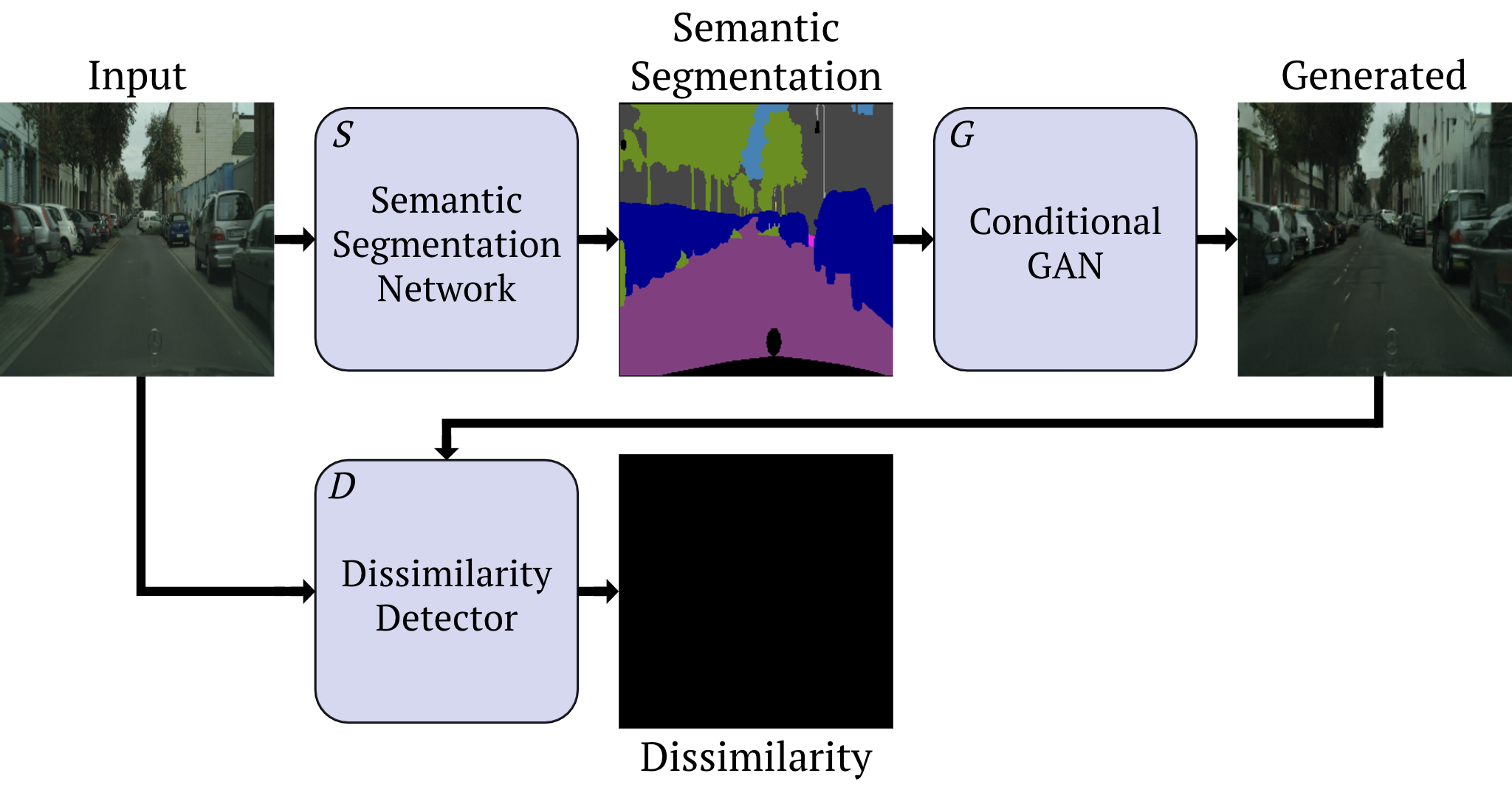}
\end{center}
   \caption{Overview of the pipeline showing the input that is processed by the semantic segmentation network \textit{S}, the resulting intermediate semantic segmentation that is further fed into a conditional GAN \textit{G} to output a synthetically generated reconstruction of the input image. The dissimilarity detector \textit{D} estimates a dissimilarity score based upon the input and the generated image.}
\label{fig:shortGraph}
\end{figure*}
\section{Related Work}

\subsection{Novelty Detection}
Novelty detection can be described as the task of differentiating between data from the same distribution as the training data, i.e. the same classes in same visual appearance, and outliers the network has neither seen nor would be able to generalize towards during training. Pimentel~\etal~\cite{pimentel2014review} categorize novelty detection methods into five general approaches: (1) The probabilistic approach, utilizing for instance a density estimation of the training data, where samples from low density areas have a low probability of originating from the training distribution. (2) Distance based approaches where the premise states that in distribution samples are closer to each other than OoD ones. (3) The reconstruction based approaches~\cite{richter2017safe,pidhorskyi2018generative,wang2018generative, schlegl2017unsupervised} use a model that is trained to reconstruct images. In distribution samples induce a smaller reconstruction error than OoD samples. (4) Domain based methods~\cite{lee2017training} define a domain where a boundary can be determined around the in-distribution samples to separate them from the the OoD samples. (5) Information-theoretic approaches utilize measures such as entropy over all samples and monitor the change in entropy after excluding samples. The idea is that excluding in distribution samples would increase entropy, while excluding OoD samples would decrease entropy.

Another approach to novelty detection includes estimating uncertainty by analyzing entropy of the predictions. Hendrycks and Gimpel~\cite{hendrycks2016baseline} showed that the simple baseline of maximum softmax probability can be used to asses uncertainty. Gal~\cite{gal2016uncertainty} shows how Bayesian deep learning can be leveraged to estimate model and input based uncertainties and introduces Monte-Carlo (MC) Dropout as an approximation. The predictive entropy is defined over the entropy of the softmax distribution of the prediction. A more sophisticated approach involves utilizing Bayesian convolutional neural networks~\cite{kendall2015bayesian} to produce a probabilistic output. The outputs are generated by sampling the posterior distribution and using the mean as the prediction and the variance as the uncertainty. The drawbacks are the need for sampling which can be a performance issue. For all these previously mentioned methods based on model uncertainty it is very difficult to model uncertainty for data that is by definition not part of the training set.

With the introduction of GANs~\cite{goodfellow2014generative}, new possibilities arose for novelty detection. In the work done by Lee~\etal~\cite{lee2017training} a GAN is trained to generate OoD samples on the boundary region around the in distribution samples to increase the performance of the classifier. Similarly, different works~\cite{wang2018generative, schlegl2017unsupervised} utilize GANs to quantify an anomaly score to assess between in and out-of-distribution samples.

Lastly, work done by Pham~\etal~\cite{pham2018bayesian} utilizes a combination of object and geometric boundary detectors to first segment all regions in the image. Then, the segmented regions are iteratively fused, using the information gained from an object detector. OoD classes can be detected as mismatches between the geometric segmentation and the object detector network.

\subsection{Image Similarity} %
Traditional methods to assess image-wise similarity use hand-crafted descriptors like Harris Corner~\cite{harris1988combined} or SIFT~\cite{lowe2004distinctive}. Firstly, important pixels are distinguished and then matched using the descriptor. With recent developments in deep learning, more methods use learned features from convolutional networks~\cite{zagoruyko2015learning,han2015matchnet,wang2014learning}. Work done by Zagoruyko and Komodakis~\cite{zagoruyko2015learning} analysed various network structures for learning feature representations and different architectures to assess general patch-wise similarity. Han \etal~\cite{han2015matchnet} present an approach with a unified feature and metric learning for similarity between patches. The similarity learned quantifies the probability if two patches were part of the same image or not. They use a feature extraction network, followed by a metric network comprised of fully connected layers to obtain a prediction. \cite{wang2014learning} cover the task of feature extraction and category-level image similarity. In that scenario, images are clustered by categories and the network needs to be able to extract important features to match images coming from the same category. To achieve this, they propose an efficient triplet sampling algorithm, which allows the network to consider both global visual properties and image semantics.

Another approach by Rippel~\etal~\cite{rippel2015metric} focuses on utilizing magnet loss rather than triplet loss to avoid the deficit of local similarity between classes. One drawback of triplet loss is that each triplet is processed independently, which doesn't take the distribution of the related class into account. Therefore, they propose the use of a magnet loss, which considers multiple samples of the class and shapes the representation space in such a way, that two close samples from different classes will have shared structural similarity, while maintaining correct separations of classes.

\section{Method}
As depicted in Figure \ref{fig:shortGraph}, the initial input image is processed by a semantic segmentation network \textit{S} producing the corresponding semantic segmentations. A generative model \textit{G}, in this case a cGAN~\cite{pix2pix2016}, generates a synthetic RGB image from the intermediate semantic segmentation. A convolutional dissimilarity detector \textit{D} then compares the original input with the generated RGB image and detects inconsistencies in the reconstruction.

\begin{figure*}
\begin{center}
\includegraphics[width=0.8\linewidth]{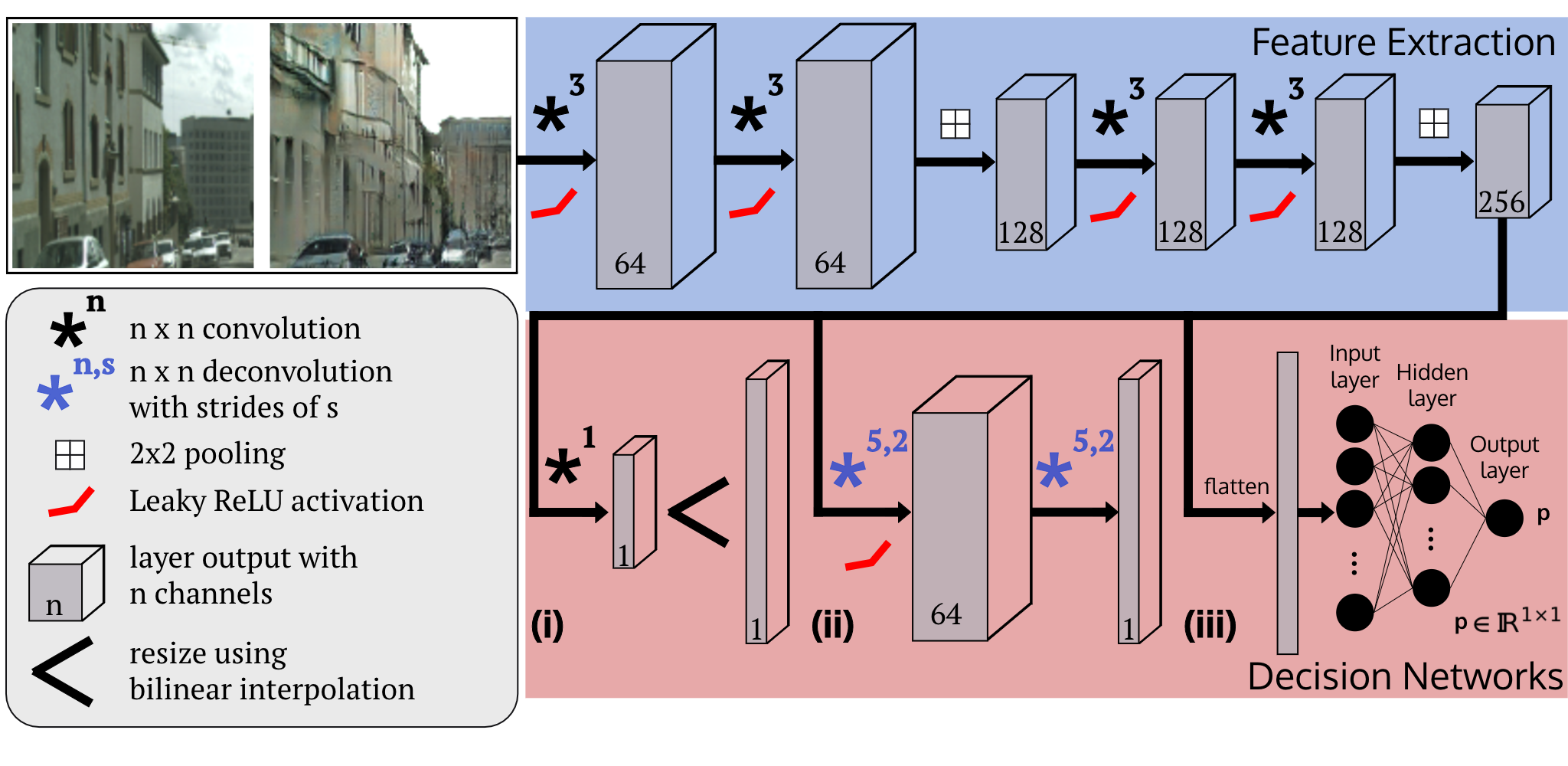}
\end{center}
   \caption{The complete network structure with the feature extractor based on \cite{simonyan2014very}. The three architectures compared in this work are shown in the red section. Method (i) uses a 1x1 convolution followed by a bilinear interpolation to scale back to patch size. Method (ii) instead uses trainable deconvolutions to scale. Lastly, method (iii) uses fully connected layers and outputs a single scalar prediction for the complete patch.}
\label{fig:networkDia}
\end{figure*}

\subsection{Semantic Segmentation and cGAN}
The semantic segmentation network \textit{S} and the generator \textit{G} are trained on the same dataset, such that within the known domain of \textit{S} we can also expect good generated images.
For semantic segmentation, we use two existing architectures, Adapnet~\cite{valada2017adapnet} and DeepLabv3+~\cite{chen2018encoder}. As previously mentioned, we use the pix2pix framework~\cite{pix2pix2016} to generate synthetic images from semantic segmentations. For the semantic segmentation network \textit{S} we are not bound to the two architectures above. Any semantic segmentation method could be used. The same applies to the generator {G}.

\subsection{Detector Objective}
 To distinguish erroneous classifications, the input image needs to be compared for inconsistencies that occurred during the reconstruction. The discriminator of the cGAN would present itself as the obvious choice, as it learned distinguishable features during training. However, the features were previously used to distinguish between real training images and generated synthetic images. In this case, real and synthetic images need to be compared. Therefore, we train the dissimilarity detector to identify the dissimilarity between real and synthetic RGB images and detect erroneous classifications by training the network on triplets~\cite{wang2014learning}. A triplet consists of three images that are passed to the network. The first and second image are paired and considered the positive example $p^+$, while the pair of the first and third image is described as the negative example $p^-$. Due to the importance of choosing appropriate negative examples, a later section is solely dedicated to explaining the process of sampling. As the goal is to detect local dissimilarities, the images are split into smaller patches to retain locality. Given two image pairs from a triplet $t_i$, the objective of the dissimilarity detector \textit{D} can be defined as
\begin{equation}
\mathcal{L}(D) = \lambda_{D}\mathbb{E}_{t_i}[log D(p^+_i)] + \mathbb{E}_{t_i}[log (1-D(p^-_i))].
\label{eq:lossfunc}
\end{equation}
This function is minimized to obtain an optimal dissimilarity detector $D$. In a later section, we analyze how the weighting parameter $\lambda_{D}$ can improve performance depending on the problem.

\subsection{Detector architecture}
The structure of the dissimilarity detector can be split into two parts similar to an encoder-decoder generator architecture~\cite{pix2pix2016}. The first part is a feature extractor with the architecture based on the first seven layers of the VGG16~\cite{simonyan2014very} network. This can be seen in the blue section of Figure \ref{fig:networkDia}. Seven layers were chosen due to the constraint that the receptive field should be limited such that locality is retained and to be equal or smaller than the GAN patch size during training. The second part can be seen as a decision network. The features extracted are processed and the decision network outputs a prediction. For the decision network three different variants were considered and are shown in the red section of Figure \ref{fig:networkDia}. The first variant (i) uses 1x1 convolutions to reduce the number of filters, then scales back by resizing using bilinear interpolation. The second approach (ii) uses multiple deconvolutions to up scale back to input size. This method has more capacity and is similar to the approach used in the generator of~\cite{pix2pix2016}. Both of these methods output a pixelwise prediction. Finally, the last method (iii) uses fully-connected layers to output a single scalar as a prediction for the complete patch. This approach was based on the metric network by \cite{han2015matchnet}.

\subsection{Sampling Strategy}
Each triplet requires three image patches. Assembling the positive pair is fairly straight forward. Given a real input image and the corresponding synthetic image, both images are split up into equally sized, non-overlapping patches. For each pair of these patches, we need to find a negative patch to complete the triplet. This patch has to be a clear negative example to reduce the noise of the training data, but should also be close enough to the positive in order for the triplet loss to work.

The negative patch needs to originate from a synthetic image, but needs to be semantically different, in the sense that if the real and synthetic patches depict a road, it would not resemble visual difference in the sense that we defined above as the task of the dissimilarity detector.. To satisfy this constraint, a negative patch is extracted from another synthetic image in the dataset. The semantics of the positive and negative patch are compared based on their respective semantic segmentations. Each pixel in the patch is compared by class affiliation. Only if a certain amount of pixels differ in class affiliation, the negative patch is accepted. If that is not the case, implying too many pixels were semantically speaking equal, a new candidate patch is chosen. To further facilitate training, the negative patch is also randomly augmented in terms of brightness, contrast and by horizontally and vertically flipping the patch.

\section{Experiments}
All experiments were conducted with Tensorflow implementations. The generative models used in the experiments were an adapted cGAN model called pix2pix~\cite{pix2pix2016} ported to Tensorflow\footnote{https://github.com/yenchenlin/pix2pix-tensorflow}. Both the Adapnet~\cite{valada2017adapnet}, DeepLabv3+~\cite{chen2018encoder} and the cGAN can be trained beforehand independently. Both are trained on the Cityscapes dataset~\cite{cordts2016cityscapes}. This dataset depicts urban street scenes over multiple European cities with corresponding ground truth semantic segmentation. The optimization of the networks is done using the Adam solver~\cite{kingma2014adam} with default parameters.
\begin{center}
 \begin{figure}
 \includegraphics[width=1.05\linewidth]{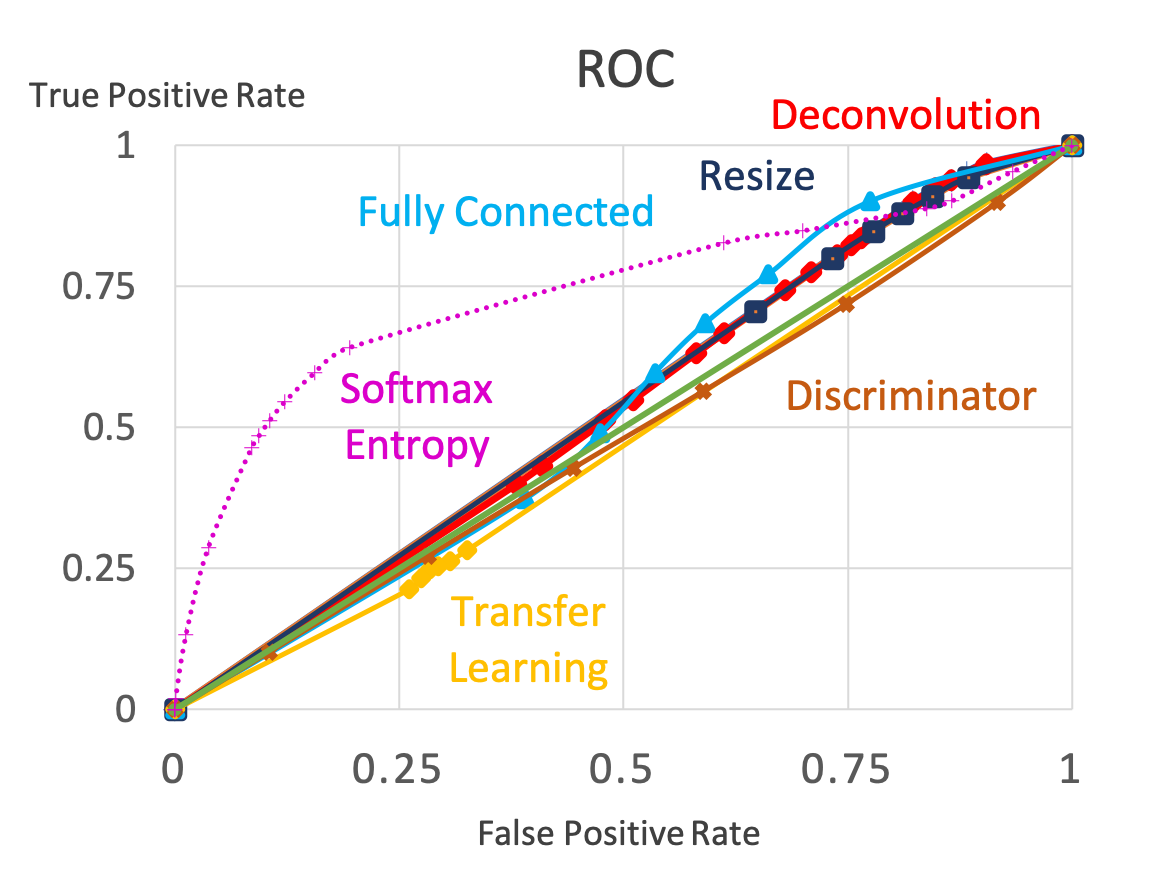}
 \caption{ROC curve for (i) Resize, (ii) Deconvolution, (iii) Fully-Connected, (iv) Transfer Learning and (v) Discriminator on a subset of the Vista dataset~\cite{neuhold2017mapillary}.}
 \label{fig:mapiROC}
 \end{figure}
\end{center}
\subsection{Error Detection}
\label{sec:ED}
\begin{figure*}[p]
\begin{center}
\includegraphics[width=1\linewidth]{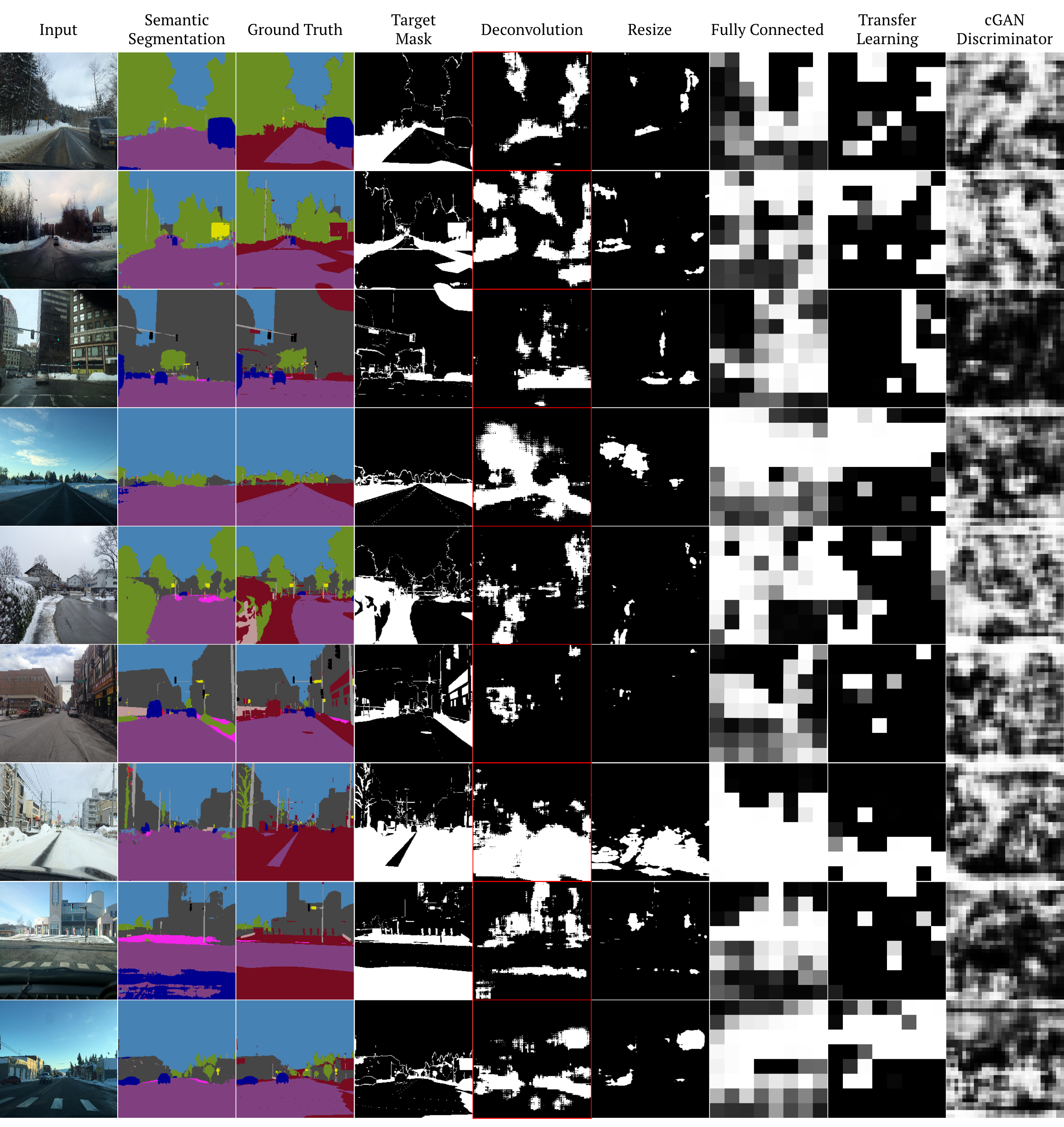}
\end{center}
   \caption{Qualitative Examples from the Error Detection experiment. The Fully Connected method seems to perform better for misclassification error, while the Deconvolution and Resize method appear to work better to detect OoD instances.}
\label{fig:quantmapi}
\end{figure*}

For this experiment, we want to assess the performance of our model to detect OoD instances as well as misclassifications introduced by semantic segmentation network on a similar dataset to Cityscapes with more classes that do not appear during training. The Mapillary Vistas dataset~\cite{neuhold2017mapillary} seemed the most suitable, as the images contained road and urban images like Cityscapes, but include other classes as well. We target all images in the dataset containing the class "boat" or "snow" and extract 100 images to evaluate. We use the DeepLabv3+~\cite{chen2018encoder} as the semantic segmentation network to achieve high quality semantic segmentation to reduce the amount of misclassifications. The objective of this task is to detect OoD classes and misclassifications. A mask is generated based off of all classes that do not appear in the Cityscapes dataset and a second mask marking any misclassifications in the semantic segmentation. Additionally, we introduce two more methods against our 3 original approaches for comparison. The fourth method (iv) employs transfer learning~\cite{chen2017photographic} by utilizing the first three layers of the pretrained discriminator of the cGAN to extract the features from the images and also uses a fully connected decision network. The fifth (v) method uses the complete pretrained discriminator of the cGAN. We evaluate the performance of all 5 methods by computing the receiver operating characteristic curve (ROC)~\cite{bradley1997use} against a softmax entropy baseline~\cite{hendrycks2016baseline}. The result can be seen in Figure \ref{fig:mapiROC}. The corresponding area under the curve (AUC) results can be found in Table \ref{tab:mapiTable}. In general it seems, that the networks which learned features performed better than those, that used a pretrained network. Both the Resize and Deconvolution method perform equally, while the fully-connected approach performs slightly better. The baseline performs well due to the misclassification errors present. For that reason the true positive rate is rather low, while keeping a very low false positive rate.
Figure \ref{fig:quantmapi} presents some qualitative results comparing all methods introduced previously.
\begin{center}
 \begin{figure}[!b]
 \includegraphics[width=1.05\linewidth]{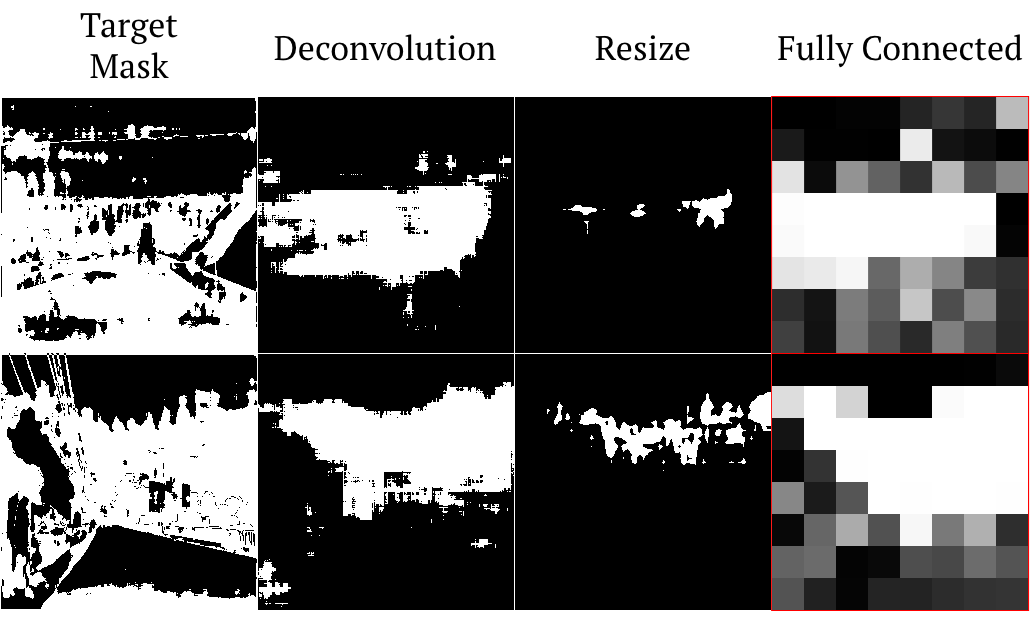}
 \caption{Dissimilarity outputs for the Deconvolution, Resize and Fully Connected approaches on the Wilddash dataset \cite{zendel2018wilddash}.}
 \label{fig:qualWild}
 \end{figure}
\end{center}
\subsection{Misclassification Detection}
\label{sec:misDet}
After looking at OoD detection combined with misclassification, we evaluate the performance of the three decision approaches only for misclassification detection. We evaluate on the Wilddash~\cite{zendel2018wilddash} dataset. This dataset uses the same classes as Cityscapes~\cite{cordts2016cityscapes} but includes road images from around the world. Additionally, some of the images contain bad weather conditions, such as foggy, rainy or snowy scenarios. Due to these conditions, it is to be expected that more misclassification error occur. Furthermore, we use AdapNet~\cite{valada2017adapnet} for our semantic segmentation network and train it on square images, as this leads to worse performance due to loss of resolution. Figure \ref{fig:qualWild} illustrates related qualitative results. Due to the abundance of OoD pixels the Fully Connected approach shows promising results. The Deconvolution method also detects certain misclassification well, but outputs very little gray areas. The Resize method struggles most. We assume that the lower capacity of the network is a likely cause for this outcome.
\begin{table}[t]
\begin{center}
\begin{tabular}{ l c }
\thickhline
Method & AUC Score \\
\hline
Deconvolution & 0.5469 \\
Resize & 0.5466 \\
Fully Connected & 0.5051 \\
Transfer Learning & 0.4458 \\
Discriminator & 0.4617 \\
\hline
Softmax Entropy & 0.7256 \\
\thickhline
\end{tabular}
\end{center}
\caption{Detection approaches are compared on a subset of the Mapillary Vistas dataset. \cite{neuhold2017mapillary}}
\label{tab:mapiTable}
\end{table}
\subsection{Loss weighting}
\begin{center}
 \begin{figure}
 \includegraphics[width=1\linewidth]{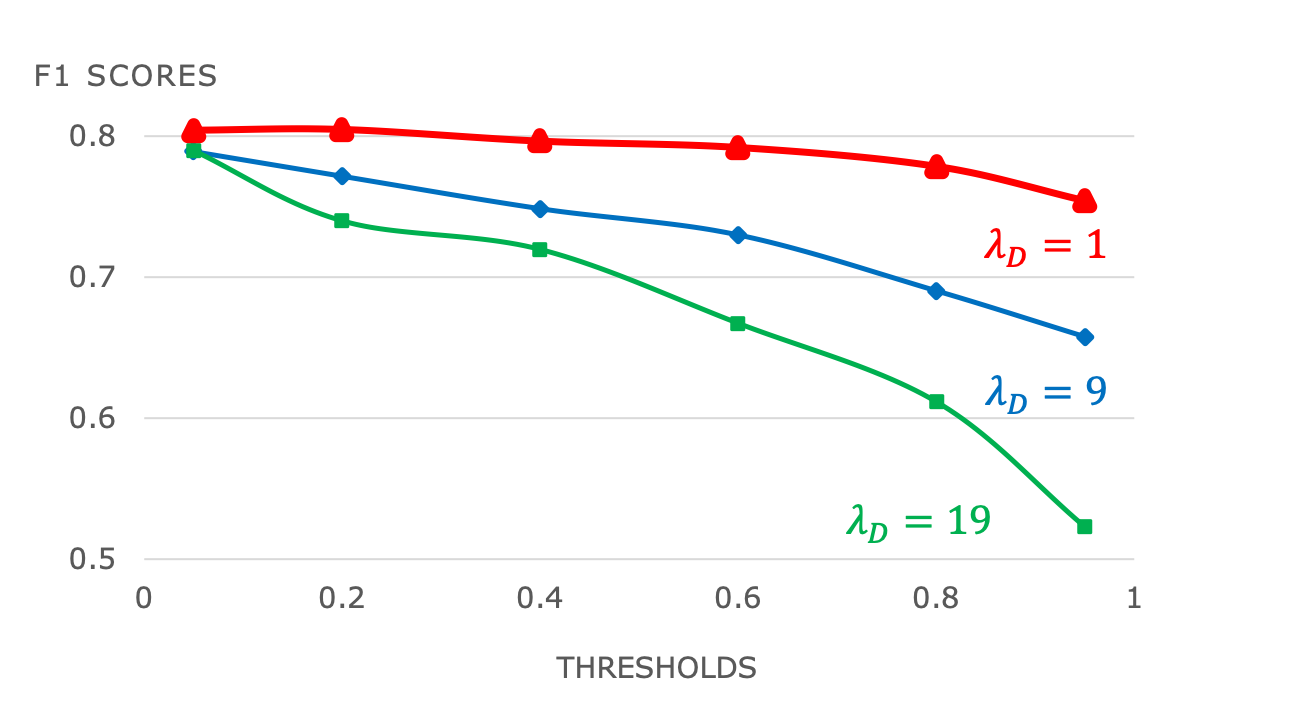}
 \caption{F1 score performance plot of the Fully Connected approach trained with three different values for $\lambda_D$.}
 \label{fig:lossWeight}
 \end{figure}
\end{center}

During training the loss function in Equation \ref{eq:lossfunc} is minimized to obtain the optimal dissimilarity detector. Depending on the task, it is possible that detecting dissimilarity is more important than detecting similarity. Depending on the ratio of in-distribution to out-of-distribution pixels, it may be beneficial to set the parameter $\lambda_D$ appropriately.
During this experiment, we analyze how the performance of the Fully Connected approach varies on the previous task explained in section \ref{sec:misDet} depending on the weighting parameter $\lambda_D$ that is specified during training. Performance is evaluated using F1 Score. The results can be found in Figure \ref{fig:lossWeight}. Due to the balanced occurrence of in and out-of-distribution pixels the performance for the Fully Connected approach produces the best results when the loss terms are equally balanced.

\section{Discussion}
Initial results indicate that generative adversarial networks combined with a learned metric network can be a useful tool for error detection and the generalization problem in semantic segmentation. From the experiments, we found that the Deconvolution approach is more effective in OoD detection, while the Fully connected approach performs better for detection of misclassification. The Deconvolution method outperforms the Resize method over all experiments due to the higher capacity in the networks.

One of the main drawbacks of the cGAN used in this work is the low resolution of images. Additionally, the training of the cGAN proved to be unstable and convergence was not reached. This loss in image quality of the generated images impairs the performance of the dissimilarity detector. Another drawback is the susceptibility of the dissimilarity detector to brightness. This causes slight differences in brightness between the input and synthetic images to always trigger a detection, even when the semantics are similar. This leads to a very high rate of false positives that deteriorates the overall performance.

\section{Conclusion}
In this paper, we introduce a novel method for error detection of semantic segmentation networks based on visual dissimilarity and evaluate various metric learning approaches to quantify dissimilarity between real and synthetic images. Our dissimilarity detector uses generative adversarial networks (GAN) to detect inconsistencies in the reconstruction. In future work, the use of other conditional GAN approaches can further improve general performance. The high false positive rate also needs further investigation, as that is the main cause for poor performance.
To conclude, our novel method could be used to extend any semantic segmentation network to detect out-of-distribution (OoD) instances. Even though the true positive rate for OoD detection in our tests showed promising results, the number of false positives was too high. This method definitely shows promise, but will require further investigation.

{\small
\bibliographystyle{ieee}
\bibliography{egbib}
}

\end{document}